\documentclass[10pt,twocolumn,letterpaper]{article}
\pdfoutput=1
\usepackage{cvpr}              
\usepackage{amsmath}
\usepackage{algorithm}
\usepackage{algpseudocode}
\usepackage{graphicx}
\usepackage{booktabs}
\usepackage{multirow}
\usepackage[table,xcdraw]{xcolor}
\usepackage{makecell}
\usepackage{stfloats}
\usepackage{pifont}
\usepackage{bm}
\usepackage{amssymb}
\usepackage{colortbl}
\usepackage{hhline}
\usepackage{boldline}
\usepackage{siunitx}
\usepackage{tabularx} 
\usepackage{graphicx}  
\usepackage{caption}   
\usepackage{subcaption} 
\usepackage{booktabs} 
\usepackage{nicematrix}   
\usepackage[accsupp]{axessibility}
\definecolor{mygreen}{RGB}{40, 140, 80} 
\definecolor{mygray}{RGB}{230,230,230}
\definecolor{blue}{RGB}{229,233,243}
\definecolor{bluegray}{RGB}{241,244,250}
\definecolor{orangegray}{RGB}{251,244,242}
\definecolor{orange}{RGB}{248,231,228}
\definecolor{gree}{RGB}{226,239,213}
\definecolor{greegray}{RGB}{245,249,241}
\definecolor{tab1huang}{RGB}{255,240,193}
\definecolor{myblue}{RGB}{196,214,230}
\definecolor{mylightblue}{RGB}{53,160,219} 
\definecolor{myllblue}{RGB}{237,242,241}
\definecolor{mediumorangegray}{RGB}{242,232,229} 
\definecolor{lightorangegray}{RGB}{251,244,242}  

\definecolor{cvprblue}{rgb}{0.21,0.49,0.74}
\usepackage[pagebackref,breaklinks,colorlinks,allcolors=cvprblue]{hyperref}

\def\paperID{33903} 
\def\confName{CVPR}
\def\confYear{2026}

\title{FedBPrompt: Federated Domain Generalization Person Re-Identification via Body Distribution Aware Visual Prompts}

\author{
Xin Xu\textsuperscript{1} \quad 
Weilong Li\textsuperscript{1} \quad 
Wei Liu\textsuperscript{1}\textsuperscript{*} \quad 
Wenke Huang\textsuperscript{3} \\ 
Zhixi Yu\textsuperscript{1} \quad 
Bin Yang\textsuperscript{2} \quad 
Xiaoying Liao\textsuperscript{4, 5}\textsuperscript{*} \quad 
Kui Jiang\textsuperscript{6} \\  
\textsuperscript{1}School of Computer Science and Technology, Wuhan University of Science and Technology, China \\
\textsuperscript{2}NERC for Multimedia Software, School of Computer Science, Wuhan University, China \\
\textsuperscript{3}Nanyang Technological University, Singapore \\
\textsuperscript{4}Changsha Bus Group, China \textsuperscript{5}Central South University of Forestry and Technology, China \\
\textsuperscript{6}Harbin Institute of Technology Zhengzhou Research Institute, China \\
\begingroup
\renewcommand\thefootnote{}\footnote{* Corresponding authors: liuwei@wust.edu.cn; wenkehuang0901@gmail.com}
\addtocounter{footnote}{-1} 
\endgroup
{\tt\small \{xuxin, liweilong, liuwei, yuzhixi\}@wust.edu.cn, wenkehuang0901@gmail.com} \\ 
{\tt\small yangbin\_cv@whu.edu.cn, kul660@psu.edu, kuijiang@whu.edu.cn} 
}

\begin{document}
\maketitle
\begingroup
\renewcommand\thefootnote{} 
\footnotetext{* Corresponding authors} 
\endgroup
\begin{abstract}

Federated Domain Generalization for Person Re-Identification (FedDG-ReID) learns domain-invariant representations from decentralized data. While Vision Transformer (ViT) is widely adopted, its global attention often fails to distinguish pedestrians from high similarity backgrounds or diverse viewpoints---a challenge amplified by cross-client distribution shifts in FedDG-ReID. 
To address this, we propose \textbf{Federated Body Distribution Aware Visual Prompt (FedBPrompt)}, introducing learnable visual prompts to guide Transformer attention toward pedestrian-centric regions. FedBPrompt employs a \textbf{Body Distribution Aware Visual Prompts Mechanism (BAPM)} comprising: \textbf{Holistic Full Body Prompts} to suppress cross-client background noise, and \textbf{Body Part Alignment Prompts} to capture fine-grained details robust to pose and viewpoint variations. 
To mitigate high communication costs, we design a \textbf{Prompt-based Fine-Tuning Strategy (PFTS)} that freezes the ViT backbone and updates only lightweight prompts, significantly reducing communication overhead while maintaining adaptability. 
Extensive experiments demonstrate that \textbf{BAPM} effectively enhances feature discrimination and cross-domain generalization, while \textbf{PFTS} achieves notable performance gains within only a few aggregation rounds. Moreover, both \textbf{BAPM} and \textbf{PFTS} can be easily integrated into existing ViT-based FedDG-ReID frameworks, making \textbf{FedBPrompt} a flexible and effective solution for federated person re-identification. The code is available at \href{https://github.com/leavlong/FedBPrompt}{https://github.com/leavlong/FedBPrompt}.





\end{abstract}    
\begin{figure}[h!]
    \centering
    \includegraphics[width=\linewidth]{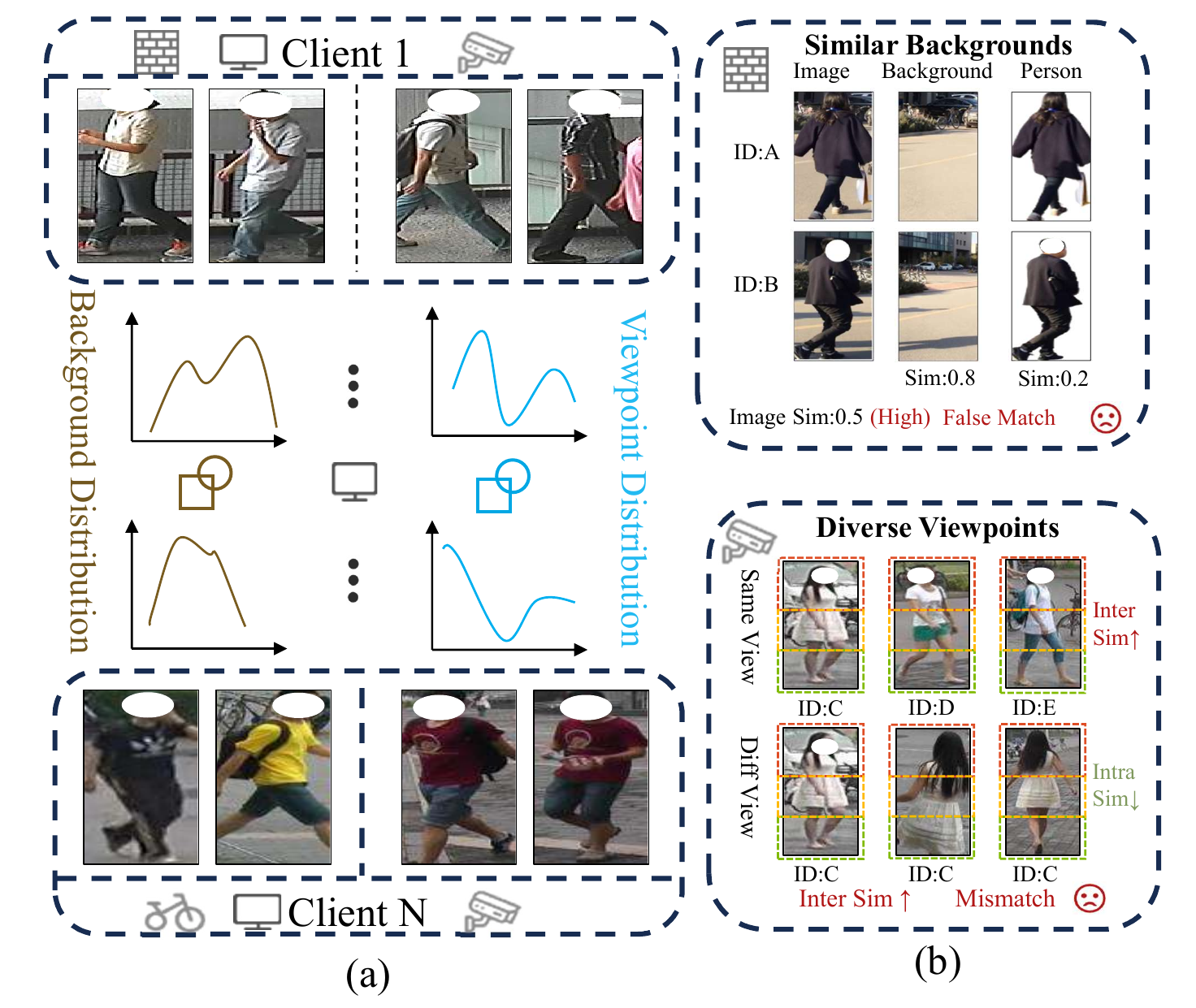}
    \caption{
        \textbf{The Dual Challenges of Client Heterogeneity in Federated Person Re-ID.} (a) In the FedDG-ReID setting, clients exhibit heterogeneous background and viewpoint distributions. (b) This data heterogeneity in turn leads to two critical failure modes: (Top) The model becomes easily distracted by dominant yet irrelevant backgrounds, causing false matches between different individuals. (Bottom) Diverse viewpoints severely misalign body parts of the same individual, which drastically reduces their feature similarity and results in mismatches.
    }
    \label{graph1}
\end{figure}

\section{Introduction}
\label{sec:intro}
Person re-identification (ReID)~\cite{luo, reid1, reid2} is a critical task in intelligent security deployment and smart city construction, which aims to accurately retrieve images of a specified pedestrian across different cameras. The advancement of deep learning has significantly propelled the development of Person ReID~\cite{reid4, reid5, msmt17, reid7}. However, limited by data acquisition challenges and real-world complexities, these methods often fail to generalize well to unseen domains. Consequently, existing methods perform poorly when affected by domain gaps. To address this problem, researchers have initiated the study of Domain Generalization Person Re-identification (DG-ReID)~\cite{snr, ni2023part, dgreid1, dgreid2}, which aims to mitigate the impact of domain gaps on model performance, enabling models trained on source domains to perform well on unseen target domains. However, existing DG-ReID methods typically require pooling different datasets for joint training, but data from cameras of different institutions cannot be centrally used due to privacy and regulatory reasons.
    

Federated Learning (FL)~\cite{fl1, fl2, feddg1, feddg2} provides a privacy-preserving framework where multiple clients can collaboratively train a model without exchanging their local data. This has given rise to the challenging yet practical field of Federated Domain Generalization for Re-ID (FedDG-ReID)~\cite{wu2021decentralised, dacs, yang2022federated, sscu, wei, xu2026fedarks}. In this domain, the Vision Transformer (ViT)~\cite{vit} has become the mainstream backbone architecture, prized for its superior performance over CNNs. However, ViT-based models suffer from a key limitation: their global attention struggles to handle pedestrians effectively, especially when dealing with images that feature challenging viewpoint changes or highly similar backgrounds. As shown in Figure~\ref{graph1}, these two challenges are exacerbated in the FedDG-ReID, where background and viewpoint distributions differ significantly across clients.


To address the dual challenges of background-induced defocusing and viewpoint-induced misalignment in a unified manner, we introduce our core contribution: the \textbf{Body Distribution Aware Visual Prompts Mechanism (BAPM)}. BAPM partitions the entire prompt set into two specialized groups: The first group of prompts, Body Part Alignment Prompts, which are evenly assigned to the upper, middle, and lower body and utilizes a constrained local attention mechanism to achieve robust alignment. And the second group of Prompts, Holistic Full Body Prompts, which capture the person's overall appearance. (Detailed division method is described in Section 4.1.) Crucially, these prompts can communicate with one another, allowing the model to learn structured, part-level features within a coherent global context. This design separates BAPM from rigid methods (\textit{e.g.},~\cite{pcb, mgn}) and ensures the generation of consistent features across diverse client viewpoints.


Furthermore, the substantial size of ViT models often leads to prohibitive communication costs when updating the entire model, a scenario which is ill-suited for the often resource-constrained federated learning environments. Therefore, to significantly reduce these federated costs without compromising model performance, we further design a \textbf{P}rompt-based \textbf{F}ine-\textbf{T}uning \textbf{S}trategy (PFTS). Drawing inspiration from PromptFL~\cite{promptfl}, this strategy begins with a pre-trained, prompt-free ReID model whose backbone is subsequently held frozen. Clients then ``implant'' a lightweight, randomly initialized set of our Body Distribution-Aware Visual Prompts and are tasked with exclusively training only these parameters. Consequently, only the updates to these prompt parameters are communicated to the central server for aggregation, reducing the uploaded parameter volume to as little as a mere 1\% of the full model. This approach drastically reduces communication overhead while simultaneously retaining high performance. Indeed, our experiments demonstrate that the prompt-based fine-tuning strategy achieves notable gains in both mAP and Rank-1 metrics within just a few rounds of aggregation. 

In summary, our main contributions are as follows:

\begin{itemize}
\item We propose FedBPrompt, a novel framework for Federated Domain Generalization in Person Re-Identification. FedBPrompt introduces learnable visual prompts to explicitly guide Transformer attention toward pedestrian-centric cues, mitigating the background bias problem in ViT-based models.
\item We design the Body Distribution Aware Visual Prompts Mechanism (BAPM), a novel mechanism that explicitly tackles the person misalignment problem arising from client-level viewpoint heterogeneity. By functionally partitioning the prompt set and enabling inter-prompt communication, BAPM guides the model to learn features that are both structured and holistically coherent, ensuring consistency across diverse client data.
\item We develop a Prompt-based Fine-Tuning Strategy (PFTS) that freezes the ViT backbone and only updates lightweight prompts, substantially reducing communication overhead. Both BAPM and PFTS can be easily integrated into various ViT-based FedDG-ReID methods, leading to consistent performance improvements.


\end{itemize}

\section{Related works}
\label{sec:formatting}
\subsection{Domain Generalization Person ReID}

Domain Generalization (DG) for Person Re-identification (ReID) has recently garnered significant attention. The primary goal is to enhance a model's generalization capability on unseen target domains by training it on multiple source domains. Existing methods predominantly improve performance through three main approaches: data augmentation, feature disentanglement, and meta-learning.


Data Augmentation-based Methods focus on enriching the diversity of training data. For instance, 
Tang \textit{et al.}~\cite{crossstyle}proposed CrossNorm, which swaps channel statistics between feature maps, Similarly, Zhou \textit{et al.}\cite{mixstyle} developed MixStyle, which linearly combines them.


Feature Disentanglement-based Methods concentrate on decoupling features into domain-invariant and domain-specific components. Jin \textit{et al.}\cite{snr} proposed Instance Normalization (IN) to filter style variations, then distill and reintegrate identity-relevant information to maintain discriminability. 
Zhang \textit{et al.}\cite{zhang2022learning} leverage causal inference to better disentangle identity and domain features for stronger generalization performance on unseen data.


Meta-Learning-based Methods aim to simulate the domain generalization process during training, enabling the model to ``learn to generalize''. Ni \textit{et al.}These methods often involve learning intermediate representations that facilitate rapid adaptation~\cite{ni2022meta} or using meta-learned modules to dynamically aggregate features from domain-specific experts~\cite{dai2021generalizable}.
However these methods require the use of centralized data, with potential risks of privacy breaches.

\subsection{Federated Learning}

Federated Learning (FL) is a distributed machine learning technique that enables model optimization on decentralized data, effectively preserving data privacy. FedAvg~\cite{fedavg} was the pioneering FL algorithm, and subsequent methods such as FedProx~\cite{fedprox}, SCAFFOLD~\cite{scaffold}, and MOON~\cite{moon} were developed to address the client drift issue arising from data heterogeneity. However, these foundational methods were primarily designed for closed-set tasks where the training and testing sets share the same class labels.

Adapting FL to the open-set nature of the person Re-ID task presents unique challenges. Early work by Zhuang \textit{et al.}~\cite{fedpav} explored protocol modifications, such as only aggregating feature extractors. Subsequently, Wu \textit{et al.}~\cite{wu2021decentralised} made a foundational contribution by formally defining Federated Domain Generalization for Re-ID (FedDG-ReID), treating each source domain as a client. More recent work has shifted focus to enhancing generalization through data augmentation. For instance, Yang \textit{et al.}~\cite{dacs} designed a model to generate novel styles for local data, while in a related vein, Liu \textit{et al.}~\cite{liu2024domain} introduced a network to create diverse fictitious domains. While focused on data-level diversity, these methods fail to directly address the key distributional challenges of FedDG-ReID: background variations causing attention defocusing, and viewpoint heterogeneity leading to misalignment. Our proposed prompting framework is explicitly designed to tackle this dual challenge at the model level, with the added benefit of being highly communication-efficient.

\subsection{Prompt Learning}

Prompting~\cite{prompt}, a paradigm originating in Large Language Models (LLMs), has evolved from manually-engineered ``discrete prompts'' to learnable ``soft prompts''~\cite{li2021prefix}. These are optimized via a parameter-efficient method known as prompt tuning~\cite{ power_prompt_tuning, P-Tuning_v2}. This concept was extended to the vision domain to overcome the limitations of text-only inputs. Early explorations by Bahng \textit{et al.}~\cite{bahng2022exploring} treated visual prompts as tunable perturbations on the image periphery. A more integrated and now prevalent approach is Visual Prompt Tuning (VPT), proposed by~\cite{vpt}, which prepends learnable tokens directly into a ViT's input sequence, drastically reducing the number of fine-tunable parameters.

The lightweight nature of prompts makes them highly suitable for efficient Federated Learning (FL). However, existing prompt-based FL methods have focused exclusively on the text domain~\cite{promptfl, fedprompt}. federated Pre-trained Language Models, both by only communicating prompt updates. Our work, in contrast, focuses on how visual prompts can be architected for federated aggregation. We propose a structured prompting mechanism designed to address the unique challenges of background interference and person misalignment in the demanding context of FedDG-ReID.
\section{Methods}
\textbf{Problem Setup.} In the context of FedDG-ReID, we consider a federated system comprising $K$ clients. Each client $k \in \{1, \ldots, K\}$ holds a local dataset $D_k=\left\{\left(\mathrm{x}_{\mathrm{i}}, \mathrm{y}_{\mathrm{i}}\right)\right\}$.from a unique source domain $S_k$ . The primary goal is to leverage this distributed data to collaboratively train a global model, $g\left(\cdot ; \Theta_{g}\right)$, capable of generalizing effectively to an unseen target domain $S_t$. The learned global parameters $\Theta_g$ can either represent the entire model, $\Theta_g = \{\Theta_b, \Theta_p\}$ in full-parameter training, or exclusively the lightweight global prompts, $\Theta_g = \Theta_p$, in our efficient fine-tuning setting.
Our method is built upon the standard Federated Learning (FL) paradigm, which enables collaborative training on decentralized data without compromising privacy. Following the common practice in FedDG-ReID research, the training process unfolds in iterative rounds, each consisting of five main steps:
\begin{itemize}
    \item Step (1): Local Training: The model on the client side is optimized using a specific training strategy.
    \item Step (2) Model Upload: Specific weights are uploaded to the server.
    \item Step (3) Model Aggregation: The weights uploaded by the clients are aggregated using a specific method.
    \item Step (4) Model Redistribution: The aggregated weights are distributed back to the clients, and steps (1) through (4) are iterated until the model converges.
    \item Step (5) Testing: The final global model is tested on target domain. The core contributions of our method are primarily situated within steps (1), (2), and (4) of this process.
\end{itemize}    

\begin{figure*}[t]
    \centering
    \includegraphics[width=\textwidth]{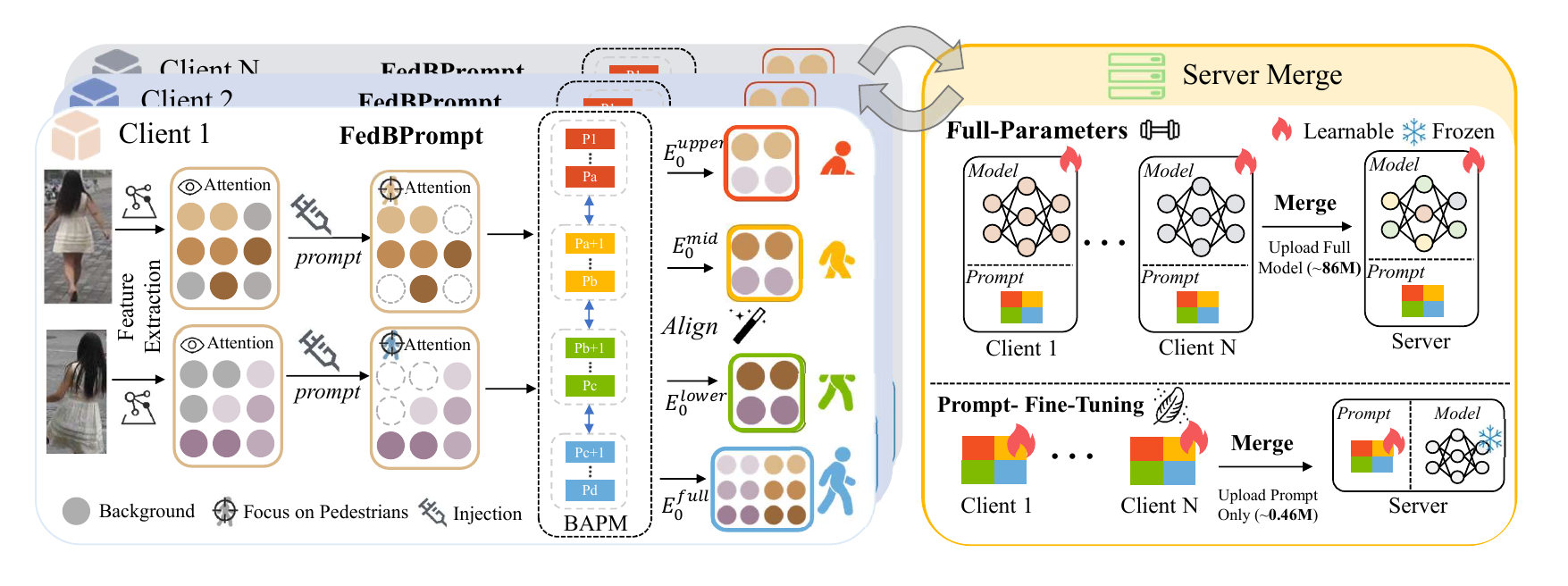}
    \vspace{-1cm}
    \caption{
        \textbf{An Overview of the Proposed FedBPrompt Framework.} 
        The framework consists of two main components.
        \textbf{(Left)} On each client, our FedBPrompt method injects learnable prompts to guide the model's attention toward pedestrian features and away from the background. The core \textbf{BAPM} then learns structured, part-level representations to solve the feature misalignment problem.
        \textbf{(Right)} The framework supports two training strategies. In \textbf{Full-Parameter} training, the entire model (\textasciitilde86M) is communicated. In contrast, our proposed \textbf{PFTS} freezes the backbone and only communicates the lightweight prompts (\textasciitilde0.46M), drastically reducing communication costs while maintaining high performance.
    }
    \vspace{-10pt}
    \label{framework}
\end{figure*}

As illustrated in Figure~\ref{framework}, the framework of our method is primarily composed of two parts: (1) full-parameter training with the introduction of the Body Distribution Aware Visual Prompts Mechanism (Section 3.1), and (2) federated prompt fine-tuning with a frozen backbone (Section 3.2). 

\subsection{Body Distribution Aware Visual Prompts}
\label{BAPM}
While excelling in FedDG-ReID, Vision Transformers (ViTs) are bottlenecked by their global attention, which struggles with two core issues stemming from client heterogeneity: background-induced defocusing and viewpoint-induced misalignment. To address both issues in a unified framework, we introduce a structured visual prompting mechanism designed to simultaneously guide the model's focus and enforce structural, part-level alignment.


\noindent\textbf{Body Distribution Aware Visual Prompts (BAPM)} (Please refer to \textbf{Appendix Section.A} for implementation details figure of BAPM.) operates by embedding a collection of some learnable prompts, $\mathbf{P}$, into each Transformer layer. Formally, we define this prompt set as: $\mathbf{P}=\{\mathbf{p}^k\in\mathbb{R}^d\mid k\in\mathbb{N},1\leq k\leq m\}.$

In our model, each layer $L_{i}$ is endowed with its own independent set of learnable prompt parameters following this structure, which we denote as $P_{i-1}$. After processing through layer $L_{i}$, we denote the collection of updated image patch embeddings as $E_{i}$, defined as: 
$\mathbf{E}=\{\mathbf{e}_i^j\in\mathbb{R}^d\mid j\in\mathbb{N},1\leq j\leq m\}.$
These embeddings, $E_{i}$,along with the updated class token $\mathbf{x}_{i-1}\in\mathbb{R}^{d}$, serve as inputs to the subsequent $(i+1)$ layer. With these definitions, the recursive update rule for the $i$-th layer is formally expressed as:
\begin{equation} \label{eq1}[\mathbf{x}_i,_-,\mathbf{E}_i]=L_i([\mathbf{x}_{i-1},\mathbf{P}_{i-1},\mathbf{E}_{i-1}])\quad i=1,\ldots,N.\end{equation}

The key innovation of BAPM lies not in the mere insertion of these prompts, but in their strategic partitioning to simultaneously address both background interference and person misalignment. Specifically, BAPM partitions the full set of prompt tokens, $\mathbf{P}$, into four disjoint subsets, which form two functionally distinct groups:
\begin{enumerate}[1.]
\item \textbf{Body Part Alignment Prompts:} The first group is designed to tackle misalignment. This is performed by three subsets of prompts: $\mathbf{P}^{\mathrm{upper}}$, corresponding to the upper body. $\mathbf{P}^\mathrm{mid}$, corresponding to the middle body. $\mathbf{P}^\mathrm{lower}$, corresponding to the lower body.

\item \textbf{Holistic Full Body Prompts:} The second group is designed for capturing the overall appearance. This is performed by the remaining subset: $\mathbf{P}^\mathrm{Full}$.
\end{enumerate}
The prompt set is the union of these: $\mathbf{P}=\mathbf{P}^\mathrm{upper}\cup\mathbf{P}^\mathrm{mid}\cup\mathbf{P}^\mathrm{lower}\cup\mathbf{P}^\mathrm{Full}$. This partitioning scheme applies to the prompt set at each layer. Consequently, the layer-specific prompt set $\mathbf{P}_{i-1}$ used in Equation~\ref{eq1} is itself composed of four corresponding disjoint subsets: $\mathbf{P}_{i-1}^{(\mathrm{upper})}$, $\mathbf{P}_{i-1}^{(\mathrm{mid})}$, \textit{etc.}

This functional separation is enforced via a constrained body distribution mechanism. The core idea is to restrict the interaction of the prompts to only their corresponding spatial regions within the sequence of image patch tokens. To formalize this, we first define these spatial regions.
Assuming the input image is divided into $H$ rows and $W$ columns of patches, for a total of $n = H \times W$ patches flattened in row-major order, we define three patch index sets: $I^{\mathrm{upper}}$, $I^{\mathrm{mid}}$, and $I^{\mathrm{lower}}$. These correspond to the top half, central half, and bottom half of the image, respectively:
\begin{equation} \label{eq2}
\begin{aligned}
    I^{\mathrm{upper}} &= \{j \mid 1 \leq j \leq n/2\},              \\
    I^{\mathrm{mid}}   &= \{j \mid n/4 + 1 \leq j \leq 3n/4\}, \\
    I^{\mathrm{lower}} &= \{j \mid n/2 + 1 \leq j \leq n\}.
\end{aligned}
\end{equation}

Note that these index sets are not mutually exclusive. Based on these overlapping index sets, we define the three subsets of image patch tokens output by layer $i$:
\begin{equation}
\begin{aligned}
    \mathbf{E}_{i}^{\mathrm{upper}} &= \{\mathbf{e}_i^j \mid j \in I_{\mathrm{upper}}\}, \\
    \mathbf{E}_{i}^{\mathrm{mid}}   &= \{\mathbf{e}_i^j \mid j \in I_{\mathrm{mid}}\}, \\
    \mathbf{E}_{i}^{\mathrm{lower}} &= \{\mathbf{e}_i^j \mid j \in I_{\mathrm{lower}}\}.
\end{aligned}
\end{equation}
The constrained attention then ensures that prompts from $\mathbf{P}_\mathrm{upper}$ can only interact with image tokens from $\mathbf{E}_\mathrm{upper}$ and likewise for the other parts. In contrast, prompts from $\mathbf{P}_\mathrm{Full}$ can attend to all image tokens.

The \textbf{BAPM} is implemented by applying a structured attention mask, $M$, within the self-attention operation. That is, for each Body Part Alignment Prompt this design only allows it to interact with its corresponding image patches. The constrained self-attention is formally expressed by the standard attention formula, modified by an additive mask:
\begin{equation}\mathrm{Attention}(Q,K,V)=\mathrm{Softmax}\left(\frac{QK^T}{\sqrt{d_k}}+\mathrm{M}\right)V.\end{equation}

The mask $M$enforces the partitioned interaction. An element $M_{ij}$is set to $-\infty$ if a query token $q_{i}$and a key token $k_{j}$ form a non-corresponding pair, and $0$ otherwise. This is formally defined as:

\begin{equation} \label{eq:mask_simple}
M_{ij} =
\begin{cases}
  -\infty & \text{if } (q_i, k_j) \in \mathcal{C}_{\text{mismatch}} \\
  0       & \text{otherwise}
\end{cases}
,
\end{equation}
where $\mathcal{C}_\text{mismatch}$ is the set of all mismatched pairs. A pair is considered mismatched if one token is a Body Part Alignment Prompt (\textit{e.g.}, from $\mathbf{P}_{i}^{\mathrm{upper}}$) and the other is an image patch from a non-corresponding region (\textit{e.g.}, not in $\mathbf{E}_{i}^{\mathrm{upper}}$).

Crucially, \textbf{all prompts interact via self-attention}, allowing for rich communication where Body Part Alignment Prompts model structural relationships and Holistic Full Body Prompts create a coherent global context from these cues. This design ensures the model learns structured yet coherent features that are consistent across diverse client viewpoints. Formally, for any two tokens $q_i$ and $k_j$ belonging to the layer-specific prompt set $\mathrm{P}_{i-1}$, we ensure the corresponding mask value is always zero:

\begin{equation} \label{eq:prompt_interaction}
M_{ij} = 0, \quad \forall q_i, k_j \in \mathbf{P}_{i-1}.
\end{equation}

This guarantees that all prompts---both Body Part Alignment Prompts($\mathrm{P}^{\mathrm{upper}}$, \textit{etc.}) and Holistic Full Body Prompts ($\mathrm{P}^{\mathrm{full}}$) can freely attend to one another. This allows the Body Part Alignment Prompts to model structural relationships while the Holistic Full Body Prompts integrate these cues into a coherent global context.

\subsection{Prompt tuning}
While full-parameter training can achieve state-of-the-art (SOTA) performance, it also incurs significant computational and communication costs. 

\noindent\textbf{Prompt-based Fine-Tuning Strategy (PFTS)}, a parameter efficient fine-tuning method that drastically reduces the costs of federated learning without a substantial compromise in performance.

In this task, the objective is to collaboratively train a single set of global prompt parameters, $\Theta_{\mathrm{p}}$
, by leveraging the distributed data from all clients. The process adheres to the principles of Federated Averaging, which achieves this goal through iterative cycles of local client optimization and then global parameter aggregation.

\begin{algorithm}[t]
\caption{Federated Learning Framework for FedDG-ReID}
\label{alg:feddg_reid}
\small 
\begin{algorithmic}[1]

\Require
\State Clients $K$; Datasets $\{D_1, \ldots, D_K\}$
\State Hyperparameters: Rounds $T$, Local Epochs $E$, Learning Rate $\eta$
\State Training mode $Mode \in \{\text{Full-Parameter, Prompt-Tuning}\}$

\Statex
\Procedure{ServerExecution}{}
    \State \textbf{Initialize:}
    \If{$Mode = \text{Full-Parameter}$}
        \State Initialize global parameters $\Theta^0 = \{\Theta_b^0, \Theta_p^0\}$
    \Else 
        \State Load frozen backbone $\Theta_b$; Initialize prompts $\Theta_p^0$
        \State Set global parameters $\Theta^0 = \{\Theta_p^0\}$
    \EndIf
    
    \For{$t = 0, 1, \ldots, T-1$}
        \State Broadcast $\Theta^t$ to all clients
        \For{each client $k \in \{1, \ldots, K\}$ \textbf{in parallel}}
            \State $\Theta_k^{t+1} \leftarrow \Call{ClientUpdate}{k, \Theta^t, D_k, Mode, \Theta_b}$
        \EndFor
        \State $\Theta^{t+1} \leftarrow \sum_{k=1}^{K} \frac{|D_k|}{\sum_{j}|D_j|} \Theta_k^{t+1}$ 
    \EndFor
    \State \textbf{return} final aggregated parameters $\Theta^T$
\EndProcedure

\Statex
\Procedure{ClientUpdate}{$k, \Theta^t, D_k, Mode, E, \eta, \Theta_b$}
    \If{$Mode = \text{Full-Parameter}$}
        \State Initialize local model $\Theta_k \leftarrow \Theta^t$
        \For{$e = 1, \ldots, E$}
            \State $\Theta_k \leftarrow \Theta_k - \eta \nabla_{\Theta_k} \mathcal{L}_{\text{ReID}}(\Theta_k; D_k)$
        \EndFor
        \State \textbf{return} $\Theta_k$
    \Else 
        \State Initialize local prompts $\Theta_{p, k} \leftarrow \Theta^t$
        \For{$e = 1, \ldots, E$}
            \State $\Theta_{p, k} \leftarrow \Theta_{p, k} - \eta \nabla_{\Theta_{p, k}} \mathcal{L}_{\text{ReID}}(\Theta_b, \Theta_{p, k}; D_k)$
        \EndFor
        \State \textbf{return} $\Theta_{p, k}$
    \EndIf
\EndProcedure

\end{algorithmic}
\end{algorithm}

Initially, the server distributes a standard, pre-trained ReID generalization model with parameters $\Theta_{\mathrm{b}}$, which contains no prompt, to all clients. Each client freezes the backbone network and ``implants'' a new set of randomly initialized Visual Prompt parameters. In communication round t, each client k receives the current global prompt $\Theta_{\mathrm{prompt}}^t$.

In communication round $t$, the objective for each client $k$ is to minimize its local objective function $\mathcal{L}_k (\Theta_{\mathrm{prompt}}).$ This function is defined as the sum of losses over all samples in the local dataset $D_{k}$:
\begin{equation}
    \mathcal{L}_k(\Theta_p) = \sum_{(x,y) \in D_k} \mathcal{L}_{\text{ReID}} \left( g(x; \Theta_b, \Theta_p), y \right).
\end{equation}

The local update for the client is therefore expressed as:
\begin{equation}
    \Theta_{\text{p}, k}^{t+1} = \operatorname*{arg\,min}_{\Theta_{\text{p}}} \mathcal{L}_k(\Theta_{\text{p}}).
\end{equation}


After each client completes local training, it sends updated prompt parameters $\Theta_{\mathrm{p},k}^{t+1}$ to the server. Upon receiving the prompts from all clients, the server performs the weighted aggregation to compute the new global prompts:

\begin{equation}\Theta_{\mathrm{p}}^{t+1}=\sum_{k=1}^K\frac{|D_k|}{\sum_{j=1}^K|D_j|}\Theta_{\mathrm{p},k}^{t+1}.\end{equation}

Finally, the server distributes the newly aggregated global prompt parameters $ \Theta_{\mathrm{prompt},k}^{t+1}$ back to the clients for the next round. The clients update their local prompts, and this process is repeated until convergence. 

This complete collaborative training procedure, which unifies both the full-parameter training and the efficient prompt-tuning method, is formally detailed in Algorithm~\ref{alg:feddg_reid}.

\section{Experiments}
\subsection{Experimental Settings}

\begin{table*}[th!]
\centering
\footnotesize 
\renewcommand{\arraystretch}{1.3} 
\setlength{\tabcolsep}{3.5pt} 

\begin{tabular}{c|c|c||cc|cc|cc|cc}
\Xhline{1.5pt}
\rowcolor{mediumorangegray}
& & & \multicolumn{2}{c|}{\textbf{MS+C2+C3→M}} & 
\multicolumn{2}{c|}{\textbf{MS+C2+M→C3}} & \multicolumn{2}{c|}{\textbf{C2+C3+M→MS}} &
\multicolumn{2}{c}{\textbf{Average}}
\\
\cline{4-11}
\multirow{-2}{*}{\textbf{Model}}\cellcolor{mediumorangegray}& \multirow{-2}{*}{\textbf{Reference}}\cellcolor{mediumorangegray}& \multirow{-2}{*}{\textbf{Method}}\cellcolor{mediumorangegray}&
\cellcolor{mediumorangegray} mAP & \cellcolor{mediumorangegray} rank-1 &
\cellcolor{mediumorangegray} mAP & \cellcolor{mediumorangegray} rank-1 &
\cellcolor{mediumorangegray} mAP & \cellcolor{mediumorangegray} rank-1 &
\cellcolor{mediumorangegray} mAP & \cellcolor{mediumorangegray} rank-1 \\ 
\hline
\hline
\multirow{3}{*}{FedProx~\cite{fedprox}} & \multirow{3}{*}{MLSys 2020} & official & 33.4 & 57.6 & 24.0 & 22.4 & 11.2 & 25.3 & 22.9 & 35.1 \\
 & & \cellcolor{lightorangegray}+PFTS & \cellcolor{lightorangegray}37.9 \textcolor{mygreen}{\scriptsize($\uparrow$4.5)} & \cellcolor{lightorangegray}61.6 \textcolor{mygreen}{\scriptsize($\uparrow$4.0)} & \cellcolor{lightorangegray}28.5 \textcolor{mygreen}{\scriptsize($\uparrow$4.5)} & \cellcolor{lightorangegray}27.4 \textcolor{mygreen}{\scriptsize($\uparrow$5.0)} & \cellcolor{lightorangegray}15.7 \textcolor{mygreen}{\scriptsize($\uparrow$4.5)} & \cellcolor{lightorangegray}36.9 \textcolor{mygreen}{\scriptsize($\uparrow$11.6)} & \cellcolor{lightorangegray}27.4 \textcolor{mygreen}{\scriptsize($\uparrow$4.5)} & \cellcolor{lightorangegray}42.0 \textcolor{mygreen}{\scriptsize($\uparrow$6.9)} \\
 & & \cellcolor{mediumorangegray}+BAPM & \cellcolor{mediumorangegray}\textbf{47.3} \textcolor{mygreen}{\scriptsize($\uparrow$13.9)} & \cellcolor{mediumorangegray}\textbf{70.9} \textcolor{mygreen}{\scriptsize($\uparrow$13.3)} & \cellcolor{mediumorangegray}\textbf{33.7} \textcolor{mygreen}{\scriptsize($\uparrow$9.7)} & \cellcolor{mediumorangegray}\textbf{33.7} \textcolor{mygreen}{\scriptsize($\uparrow$11.3)} & \cellcolor{mediumorangegray}\textbf{17.7} \textcolor{mygreen}{\scriptsize($\uparrow$6.5)} & \cellcolor{mediumorangegray}\textbf{41.2} \textcolor{mygreen}{\scriptsize($\uparrow$15.9)} & \cellcolor{mediumorangegray}\textbf{32.9} \textcolor{mygreen}{\scriptsize($\uparrow$10.0)} & \cellcolor{mediumorangegray}\textbf{48.6} \textcolor{mygreen}{\scriptsize($\uparrow$13.5)} \\
\hline
\hline
\multirow{3}{*}{Mixstyle~\cite{mixstyle}} & \multirow{3}{*}{ICLR 2021} & official & 29.3 & 53.3 & 19.6 & 18.6 & 12.4 & 27.9 & 20.4 & 33.3 \\
 & & \cellcolor{lightorangegray}+PFTS & \cellcolor{lightorangegray}34.5 \textcolor{mygreen}{\scriptsize($\uparrow$5.2)} & \cellcolor{lightorangegray}58.7 \textcolor{mygreen}{\scriptsize($\uparrow$5.4)} & \cellcolor{lightorangegray}23.9 \textcolor{mygreen}{\scriptsize($\uparrow$4.3)} & \cellcolor{lightorangegray}22.9 \textcolor{mygreen}{\scriptsize($\uparrow$4.3)} & \cellcolor{lightorangegray}17.5 \textcolor{mygreen}{\scriptsize($\uparrow$5.1)} & \cellcolor{lightorangegray}40.4 \textcolor{mygreen}{\scriptsize($\uparrow$12.5)} & \cellcolor{lightorangegray}25.3 \textcolor{mygreen}{\scriptsize($\uparrow$4.9)} & \cellcolor{lightorangegray}40.7 \textcolor{mygreen}{\scriptsize($\uparrow$7.4)} \\
 & & \cellcolor{mediumorangegray}+BAPM & \cellcolor{mediumorangegray}\textbf{48.3} \textcolor{mygreen}{\scriptsize($\uparrow$19.0)} & \cellcolor{mediumorangegray}\textbf{73.1} \textcolor{mygreen}{\scriptsize($\uparrow$19.8)} & \cellcolor{mediumorangegray}\textbf{33.3} \textcolor{mygreen}{\scriptsize($\uparrow$13.7)} & \cellcolor{mediumorangegray}\textbf{32.9} \textcolor{mygreen}{\scriptsize($\uparrow$14.3)} & \cellcolor{mediumorangegray}\textbf{21.3} \textcolor{mygreen}{\scriptsize($\uparrow$8.9)} & \cellcolor{mediumorangegray}\textbf{47.1} \textcolor{mygreen}{\scriptsize($\uparrow$19.2)} & \cellcolor{mediumorangegray}\textbf{34.3} \textcolor{mygreen}{\scriptsize($\uparrow$13.9)} & \cellcolor{mediumorangegray}\textbf{51.0} \textcolor{mygreen}{\scriptsize($\uparrow$17.7)} \\
\cline{1-11}
 \multirow{3}{*}{CrossStyle~\cite{crossstyle}} & \multirow{3}{*}{ICCV 2021} & official & 33.3 & 58.5 & 25.4 & 24.4 & 12.1 & 28.2 & 23.6 & 37.0 \\
 & & \cellcolor{lightorangegray}+PFTS & \cellcolor{lightorangegray}41.4 \textcolor{mygreen}{\scriptsize($\uparrow$8.1)} & \cellcolor{lightorangegray}66.1 \textcolor{mygreen}{\scriptsize($\uparrow$7.6)} & \cellcolor{lightorangegray}29.7 \textcolor{mygreen}{\scriptsize($\uparrow$4.3)} & \cellcolor{lightorangegray}29.3 \textcolor{mygreen}{\scriptsize($\uparrow$4.9)} & \cellcolor{lightorangegray}20.6 \textcolor{mygreen}{\scriptsize($\uparrow$8.5)} & \cellcolor{lightorangegray}46.4 \textcolor{mygreen}{\scriptsize($\uparrow$18.2)} & \cellcolor{lightorangegray}30.6 \textcolor{mygreen}{\scriptsize($\uparrow$7.0)} & \cellcolor{lightorangegray}47.3 \textcolor{mygreen}{\scriptsize($\uparrow$10.3)} \\
 & & \cellcolor{mediumorangegray}+BAPM & \cellcolor{mediumorangegray}\textbf{49.4} \textcolor{mygreen}{\scriptsize($\uparrow$16.1)} & \cellcolor{mediumorangegray}\textbf{73.6} \textcolor{mygreen}{\scriptsize($\uparrow$15.1)} & \cellcolor{mediumorangegray}\textbf{32.0} \textcolor{mygreen}{\scriptsize($\uparrow$6.6)} & \cellcolor{mediumorangegray}\textbf{31.6} \textcolor{mygreen}{\scriptsize($\uparrow$7.2)} & \cellcolor{mediumorangegray}\textbf{22.7} \textcolor{mygreen}{\scriptsize($\uparrow$10.6)} & \cellcolor{mediumorangegray}\textbf{49.1} \textcolor{mygreen}{\scriptsize($\uparrow$20.9)} & \cellcolor{mediumorangegray}\textbf{34.7} \textcolor{mygreen}{\scriptsize($\uparrow$11.1)} & \cellcolor{mediumorangegray}\textbf{51.4} \textcolor{mygreen}{\scriptsize($\uparrow$14.4)} \\
\hline
\hline
\multirow{3}{*}{FedReID~\cite{wu2021decentralised}} & \multirow{3}{*}{AAAI 2021} & official & 35.3 & 58.9 & 25.5 & 24.6 & 9.4 & 22.1 & 23.4 & 35.2 \\
 & & \cellcolor{lightorangegray}+PFTS & \cellcolor{lightorangegray}41.0 \textcolor{mygreen}{\scriptsize($\uparrow$5.7)} & \cellcolor{lightorangegray}65.1 \textcolor{mygreen}{\scriptsize($\uparrow$6.2)} & \cellcolor{lightorangegray}28.5 \textcolor{mygreen}{\scriptsize($\uparrow$3.0)} & \cellcolor{lightorangegray}27.4 \textcolor{mygreen}{\scriptsize($\uparrow$2.8)} & \cellcolor{lightorangegray}13.7 \textcolor{mygreen}{\scriptsize($\uparrow$4.3)} & \cellcolor{lightorangegray}33.0 \textcolor{mygreen}{\scriptsize($\uparrow$10.9)} & \cellcolor{lightorangegray}27.7 \textcolor{mygreen}{\scriptsize($\uparrow$4.3)} & \cellcolor{lightorangegray}41.8 \textcolor{mygreen}{\scriptsize($\uparrow$6.6)} \\
 & & \cellcolor{mediumorangegray}+BAPM & \cellcolor{mediumorangegray}\textbf{46.9} \textcolor{mygreen}{\scriptsize($\uparrow$11.6)} & \cellcolor{mediumorangegray}\textbf{70.8} \textcolor{mygreen}{\scriptsize($\uparrow$11.9)} & \cellcolor{mediumorangegray}\textbf{32.3} \textcolor{mygreen}{\scriptsize($\uparrow$6.8)} & \cellcolor{mediumorangegray}\textbf{31.8} \textcolor{mygreen}{\scriptsize($\uparrow$7.2)} & \cellcolor{mediumorangegray}\textbf{17.2} \textcolor{mygreen}{\scriptsize($\uparrow$7.8)} & \cellcolor{mediumorangegray}\textbf{40.1} \textcolor{mygreen}{\scriptsize($\uparrow$18.0)} & \cellcolor{mediumorangegray}\textbf{32.1} \textcolor{mygreen}{\scriptsize($\uparrow$8.7)} & \cellcolor{mediumorangegray}\textbf{47.6} \textcolor{mygreen}{\scriptsize($\uparrow$12.4)} \\
\cline{1-11}
 \multirow{3}{*}{FedPav~\cite{fedpav}} & \multirow{3}{*}{MM 2020} & official & 33.3 & 57.1 & 15.9 & 14.5 & 8.5 & 20.7 & 19.2 & 30.8 \\
 & & \cellcolor{lightorangegray}+PFTS & \cellcolor{lightorangegray}36.3 \textcolor{mygreen}{\scriptsize($\uparrow$3.0)} & \cellcolor{lightorangegray}61.0 \textcolor{mygreen}{\scriptsize($\uparrow$3.9)} & \cellcolor{lightorangegray}23.0 \textcolor{mygreen}{\scriptsize($\uparrow$7.1)} & \cellcolor{lightorangegray}22.6 \textcolor{mygreen}{\scriptsize($\uparrow$8.1)} & \cellcolor{lightorangegray}12.5 \textcolor{mygreen}{\scriptsize($\uparrow$4.0)} & \cellcolor{lightorangegray}30.8 \textcolor{mygreen}{\scriptsize($\uparrow$10.1)} & \cellcolor{lightorangegray}23.9 \textcolor{mygreen}{\scriptsize($\uparrow$4.7)} & \cellcolor{lightorangegray}38.1 \textcolor{mygreen}{\scriptsize($\uparrow$7.3)} \\
 & & \cellcolor{mediumorangegray}+BAPM & \cellcolor{mediumorangegray}\textbf{45.7} \textcolor{mygreen}{\scriptsize($\uparrow$12.4)} & \cellcolor{mediumorangegray}\textbf{71.2} \textcolor{mygreen}{\scriptsize($\uparrow$14.1)} & \cellcolor{mediumorangegray}\textbf{34.0} \textcolor{mygreen}{\scriptsize($\uparrow$18.1)} & \cellcolor{mediumorangegray}\textbf{33.9} \textcolor{mygreen}{\scriptsize($\uparrow$19.4)} & \cellcolor{mediumorangegray}\textbf{17.4} \textcolor{mygreen}{\scriptsize($\uparrow$8.9)} & \cellcolor{mediumorangegray}\textbf{40.1} \textcolor{mygreen}{\scriptsize($\uparrow$19.4)} & \cellcolor{mediumorangegray}\textbf{32.4} \textcolor{mygreen}{\scriptsize($\uparrow$13.2)} & \cellcolor{mediumorangegray}\textbf{48.4} \textcolor{mygreen}{\scriptsize($\uparrow$17.6)} \\
\hline
\hline
\multirow{3}{*}{DACS~\cite{dacs}} & \multirow{3}{*}{AAAI 2024} & official & 39.6 & 64.5 & 31.3 & 30.9 & 13.7 & 30.1 & 28.2 & 41.8 \\
 & & \cellcolor{lightorangegray}+PFTS & \cellcolor{lightorangegray}42.1 \textcolor{mygreen}{\scriptsize($\uparrow$2.5)} & \cellcolor{lightorangegray}67.1 \textcolor{mygreen}{\scriptsize($\uparrow$2.6)} & \cellcolor{lightorangegray}\textbf{35.1} \textcolor{mygreen}{\scriptsize($\uparrow$3.8)} & \cellcolor{lightorangegray}\textbf{36.2} \textcolor{mygreen}{\scriptsize($\uparrow$5.3)} & \cellcolor{lightorangegray}18.6 \textcolor{mygreen}{\scriptsize($\uparrow$4.9)} & \cellcolor{lightorangegray}42.2 \textcolor{mygreen}{\scriptsize($\uparrow$12.1)} & \cellcolor{lightorangegray}31.9 \textcolor{mygreen}{\scriptsize($\uparrow$3.7)} & \cellcolor{lightorangegray}48.5 \textcolor{mygreen}{\scriptsize($\uparrow$6.7)} \\
 & & \cellcolor{mediumorangegray}+BAPM & \cellcolor{mediumorangegray}\textbf{49.7} \textcolor{mygreen}{\scriptsize($\uparrow$10.1)} & \cellcolor{mediumorangegray}\textbf{74.3} \textcolor{mygreen}{\scriptsize($\uparrow$9.8)} & \cellcolor{mediumorangegray}34.6 \textcolor{mygreen}{\scriptsize($\uparrow$3.3)} & \cellcolor{mediumorangegray}34.8 \textcolor{mygreen}{\scriptsize($\uparrow$3.9)} & \cellcolor{mediumorangegray}\textbf{21.9} \textcolor{mygreen}{\scriptsize($\uparrow$8.2)} & \cellcolor{mediumorangegray}\textbf{48.5} \textcolor{mygreen}{\scriptsize($\uparrow$18.4)} & \cellcolor{mediumorangegray}\textbf{35.4} \textcolor{mygreen}{\scriptsize($\uparrow$7.2)} & \cellcolor{mediumorangegray}\textbf{52.5} \textcolor{mygreen}{\scriptsize($\uparrow$10.7)} \\
\cline{1-11}
 \multirow{3}{*}{SSCU~\cite{sscu}} & \multirow{3}{*}{MM 2025} & official & 46.3 & 69.6 & 33.7 & 33.4 & 20.0 & 43.7 & 33.3 & 48.9 \\
 & & \cellcolor{lightorangegray}+PFTS & \cellcolor{lightorangegray}48.9 \textcolor{mygreen}{\scriptsize($\uparrow$2.6)} & \cellcolor{lightorangegray}72.4 \textcolor{mygreen}{\scriptsize($\uparrow$2.8)} & \cellcolor{lightorangegray}35.5 \textcolor{mygreen}{\scriptsize($\uparrow$1.8)} & \cellcolor{lightorangegray}35.8 \textcolor{mygreen}{\scriptsize($\uparrow$2.4)} & \cellcolor{lightorangegray}21.3 \textcolor{mygreen}{\scriptsize($\uparrow$1.3)} & \cellcolor{lightorangegray}46.0 \textcolor{mygreen}{\scriptsize($\uparrow$2.3)} & \cellcolor{lightorangegray}35.2 \textcolor{mygreen}{\scriptsize($\uparrow$1.9)} & \cellcolor{lightorangegray}51.4 \textcolor{mygreen}{\scriptsize($\uparrow$2.5)} \\
 & & \cellcolor{mediumorangegray}+BAPM & \cellcolor{mediumorangegray}\textbf{49.1} \textcolor{mygreen}{\scriptsize($\uparrow$2.8)} & \cellcolor{mediumorangegray}\textbf{73.4} \textcolor{mygreen}{\scriptsize($\uparrow$3.8)} & \cellcolor{mediumorangegray}\textbf{37.4} \textcolor{mygreen}{\scriptsize($\uparrow$3.7)} & \cellcolor{mediumorangegray}\textbf{38.4} \textcolor{mygreen}{\scriptsize($\uparrow$5.0)} & \cellcolor{mediumorangegray}\textbf{23.4} \textcolor{mygreen}{\scriptsize($\uparrow$3.4)} & \cellcolor{mediumorangegray}\textbf{49.5} \textcolor{mygreen}{\scriptsize($\uparrow$5.8)} & \cellcolor{mediumorangegray}\textbf{36.6} \textcolor{mygreen}{\scriptsize($\uparrow$3.3)} & \cellcolor{mediumorangegray}\textbf{53.8} \textcolor{mygreen}{\scriptsize($\uparrow$4.9)} \\ 
\Xhline{1.5pt}
\end{tabular}
\caption{\textbf{Comparison of different methods under \textit{protocol-1}.} M: Market1501, C2: CUHK02, C3: CUHK03, MS: MSMT17. Average represents the average performance over three unseen domains. }
\vspace{-6mm}
\label{sota}
\end{table*}

\textbf{Datasets and Protocols.} Our experimental evaluation is conducted on four widely-used, large-scale Re-ID datasets: CUHK02 (C2)~\cite{cuhk02}, CUHK03 (C3)~\cite{cuhk03}, Market1501 (M)~\cite{market1501}, and MSMT17 (MS)~\cite{msmt17}. To thoroughly assess the model's generalization capabilities under various federated domain generalization settings, we employ two distinct evaluation protocols:

\textit{Protocol-1 (Leave-One-Out)}: This protocol follows a standard leave-one-domain-out cross-validation scheme. In each fold, one dataset is designated as the unseen target domain for testing, while the remaining three serve as the source domains for federated training. This results in four separate training/testing splits (\textit{e.g.}, training on C2+C3+M, testing on MS).

\textit{Protocol-2 (Source-Domain Performance)}: To evaluate the model's performance on the source domains themselves (\textit{i.e.}, in-domain testing), we train the model on a federation of CUHK02, CUHK03, and Market1501. The final global model is then evaluated on the individual test sets of each of these three source domains.

\textbf{Implementation Details.} \label{sec:prompts}In our federated setting, each source domain is treated as an individual client. For our implementation, we employ a ViT-B/16 as the backbone within the federated training framework established by SSCU\cite{sscu}. 
For our proposed BAPM, we use a total of 50 learnable prompt tokens. The first group, targeting misalignment, consists of 15 prompts, which are evenly divided into three subsets of 5 for the upper, middle, and lower body, respectively. The second group, capturing the overall appearance, is composed of the remaining 35 prompts. (A sensitivity analysis on the number of prompts is presented in \textbf{Appendix Section.B}.)

\subsection{Comparisons With State-of-the-art Methods}
To demonstrate the effectiveness of our method, we integrate our BAPM into several existing methods and compare their performance under two distinct training settings: full-parameter training and prompt fine-tuning.

\textit{Protocol-1}: As presented in Tabel~\ref{sota}, the methods for comparison can be divided into four main categories: (1) First, classical Federated Learning algorithms like FedProx\cite{fedprox}. (2) Second, style-based domain generalization methods, MixStyle\cite{mixstyle} and CrossStyle\cite{crossstyle}, which we deploy on the client-side to generate diverse data. (3) Next, representative federated Re-ID methods such as FedPav\cite{fedpav} and FedReid\cite{wu2021decentralised}. (4) Finally, federated domain generalization for Re-ID methods, including DACS\cite{dacs} and SSCU. The results demonstrate that our method boosts performance across all baselines in both fine-tuning (+PFTS) and full-parameter (+BAPM) settings. 
Notably, on the challenging ``M+C2+C3$\to$MS'' task, our BAPM strategy improves the strong baseline SSCU by \textbf{3.4\%} in mAP and \textbf{5.8\%} in Rank-1. 
Even for weaker baselines like FedProx, our method yields gains of \textbf{13.9\%} in mAP and \textbf{13.3\%} in Rank-1.
On average across all scenarios, our full-parameter strategy surpasses the state-of-the-art SSCU by \textbf{3.3\%} in mAP and \textbf{4.9\%} in Rank-1, validating the robustness of our approach against diverse domain shifts.

\textit{Protocol-2}: As shown in Table~\ref{tab:source}, we directly deploy the converged global models trained under the ``M+C2+C3→MS'' federated setting back to their constituent source domains. For testing, we evaluate our framework by integrating it into a diverse set of representative baselines (including FedProx, FedPav, FedReID, DACS, and SSCU) to verify its universality. The experimental results compellingly indicate that our method not only possesses outstanding cross-domain generalization performance but also simultaneously ensures optimal recognition accuracy on the original source domains across various architectural strategies.

\subsection{Ablation Study on BAPM Components.}\label{sec:ablation study} To disentangle and validate the contributions of the distinct components within our proposed BAPM, we conduct a comprehensive ablation study. We compare three model variants, using SSCU, DACS, and FedProx as a representative baseline: (Additional ablation studies on our method's components are presented in \textbf{Appendix Section.C}.) 

\begin{table}[th!]
\resizebox{\linewidth}{!}{
\centering
\begin{tabular}{c|cc|cc|cc}
\Xhline{1.5pt}
\rowcolor{myblue}
\multicolumn{1}{c|}{} & \multicolumn{2}{c|}{M+C2+C3} & \multicolumn{2}{c|}{M+C2+C3} & \multicolumn{2}{c}{M+C2+C3} \\
\rowcolor{myblue}
\multicolumn{1}{c|}{}& \multicolumn{2}{c|}{$\rightarrow$ M} & \multicolumn{2}{c|}{$\rightarrow$ C2} & \multicolumn{2}{c}{$\rightarrow$ C3} \\
\cline{2-3} \cline{4-5} \cline{6-7}
\rowcolor{myblue}
\multicolumn{1}{c|}{\cellcolor{myblue}\multirow{-3}{*}{\textbf{Methods}}} & mAP & rank-1 & mAP & rank-1 & mAP & rank-1 \\
\hline
{FedProx} & 68.7 & 85.1 & 79.9 & 79.3 & 44.6 & 45.6 \\
\rowcolor{myllblue}
{FedProx+BAPM} & \textbf{72.9} & \textbf{88.4} & \textbf{83.1} & \textbf{83.3} & \textbf{49.5} & \textbf{50.6} \\
{FedPav} & 66.5 & 83.2 & 78.5 & 78.5 & 42.3 & 42.9 \\
\rowcolor{myllblue}
{FedPav+BAPM} & \textbf{74.3} & \textbf{88.5} & \textbf{84.2} & \textbf{84.1} & \textbf{52.2} & \textbf{53.0} \\
{FedReID} & 69.2 & 85.5 & 79.2 & 77.8 & 41.7 & 42.1 \\
\rowcolor{myllblue}
{FedReID+BAPM} & \textbf{72.2} & \textbf{88.2} & \textbf{80.6} & \textbf{79.7} & \textbf{45.1} & \textbf{46.3} \\
{DACS} & 71.5 & 87.1 & 83.5 & 83.3 & 46.7 & 47.9 \\
\rowcolor{myllblue}
{DACS+BAPM} & \textbf{75.6} & \textbf{89.8} & \textbf{85.9} & \textbf{84.7} & \textbf{52.8} & \textbf{54.4} \\
{SSCU} & 76.5 & 89.8 & 86.7 & 86.0 & 53.2 & 54.6 \\
\rowcolor{myllblue}
{SSCU+BAPM} & \textbf{78.6} & \textbf{91.2} & \textbf{88.7} & \textbf{88.1} & \textbf{59.0} & \textbf{59.7} \\
\Xhline{1.5pt}
\end{tabular}}
\caption{\textbf{Comparison of different methods on \textit{Protocol-2}.} Under this protocol, we focus on the model's identification performance on the source domains, the most basic yet often overlooked capability by existing methods.}
\vspace{-4mm}
\label{tab:source}
\end{table}

\begin{enumerate}[noitemsep, topsep=0pt, parsep=0pt, partopsep=0pt]
\item \textbf{Baseline:} The original baseline model (\textit{e.g.}, SSCU) without injecting any learnable prompts.
\item \textbf{Baseline + Holistic Full Body Prompts:} Equipped only with Holistic Full Body Prompts ($\mathbf{P}^{\mathrm{full}}$) that attend to all image patches globally without spatial constraints.
\item \textbf{Baseline + Body Part Alignment Prompts:} Equipped only with the three subsets of Body Part Alignment Prompts ($\mathbf{P}^{\mathrm{upper}}$, $\mathbf{P}^{\mathrm{mid}}$, $\mathbf{P}^{\mathrm{lower}}$) for partitioned regions.
\item \textbf{Baseline + BAPM:} The implementation integrating both prompt groups via the constrained attention mechanism.
\end{enumerate}

\begin{table}[h!]
\resizebox{\linewidth}{!}{
\centering
\begin{tabular}{c|cc|cc|cc|cc}
\Xhline{1.5pt}

\rowcolor{gree}
\multicolumn{1}{c|}{\cellcolor{gree}} & \multicolumn{2}{c|}{\cellcolor{gree}\textbf{Prompt Type}} & \multicolumn{2}{c|}{\cellcolor{gree}\textbf{C2+C3+MS$\rightarrow$M}} & \multicolumn{2}{c|}{\cellcolor{gree}\textbf{C2+M+MS$\rightarrow$C3}} & \multicolumn{2}{c}{\cellcolor{gree}\textbf{C2+C3+M$\rightarrow$MS}} \\
\cline{2-3} \cline{4-5} \cline{6-7} \cline{8-9}
\rowcolor{gree}
\multicolumn{1}{c|}{\cellcolor{gree}\multirow{-2}{*}{\textbf{Methods}}} & \textbf{Global} & \textbf{Local} & mAP & rank-1 & mAP & rank-1 & mAP & rank-1 \\
\hline\hline

 & $\times$ & $\times$ & 46.3 & 69.6 & 33.7 & 33.4 & 20.0 & 43.7 \\
\rowcolor{greegray}
 & \checkmark & $\times$ & 48.4 & 73.2 & 36.7 & 36.8 & 22.9 & 48.2 \\
 & $\times$ & \checkmark & 47.7 & 71.8 & 37.5 & 37.9 & 22.7 & 48.5 \\
\rowcolor{greegray}
\multirow{-4}{*}{\textbf{SSCU}} & \checkmark & \checkmark & \textbf{49.1} & \textbf{73.4} & \textbf{37.4} & \textbf{38.4} & \textbf{23.4} & \textbf{49.5} \\
\hline

 & $\times$ & $\times$ & 39.6 & 64.5 & 32.5 & 33.0 & 13.7 & 30.1 \\
\rowcolor{greegray}
 & \checkmark & $\times$ & 48.4 & 72.8 & 34.3 & 34.2 & 17.3 & 36.4 \\
 & $\times$ & \checkmark & 48.0 & 72.3 & 34.9 & 35.0 & 20.7 & 46.5 \\
\rowcolor{greegray}
\multirow{-4}{*}{\textbf{DACS}} & \checkmark & \checkmark & \textbf{49.7} & \textbf{74.3} & \textbf{37.2} & \textbf{37.9} & \textbf{21.9} & \textbf{48.5} \\
\hline

 & $\times$ & $\times$ & 33.4 & 57.6 & 24.0 & 22.4 & 11.2 & 25.3 \\
\rowcolor{greegray}
 & \checkmark & $\times$ & 44.8 & 69.3 & 30.1 & 29.5 & 14.2 & 31.3 \\
 & $\times$ & \checkmark & 46.0 & 70.8 & 33.3 & 33.7 & 18.6 & 42.5 \\
\rowcolor{greegray}
\multirow{-4}{*}{\textbf{FedProx}} & \checkmark & \checkmark & \textbf{47.3} & \textbf{70.9} & \textbf{33.7} & \textbf{33.9} & \textbf{18.8} & \textbf{43.1} \\
\Xhline{1.5pt}
\end{tabular}%
}
\caption{\textbf{Ablation study on key components.} We validate the effectiveness of different prompt types. ``Global'' and ``Local'' denote Holistic Full Body Prompts and Body Part Alignment Prompts, respectively. Their combination forms our full BAPM.}
\label{tab:ablation}
\end{table}

The results, presented in Table~\ref{tab:ablation}, reveal a clear, incremental performance improvement. While adding holistic prompts alone already boosts the baseline performance (validating our initial motivation), the full BAPM model achieves significantly superior results. This compellingly demonstrates that the structural guidance provided by our body part alignment prompt is the critical component for tackling the feature misalignment challenge and is the primary driver of the overall performance gain.

\vspace{5mm}
\begin{figure}[!h]
    \centering
    \includegraphics[width=\linewidth]{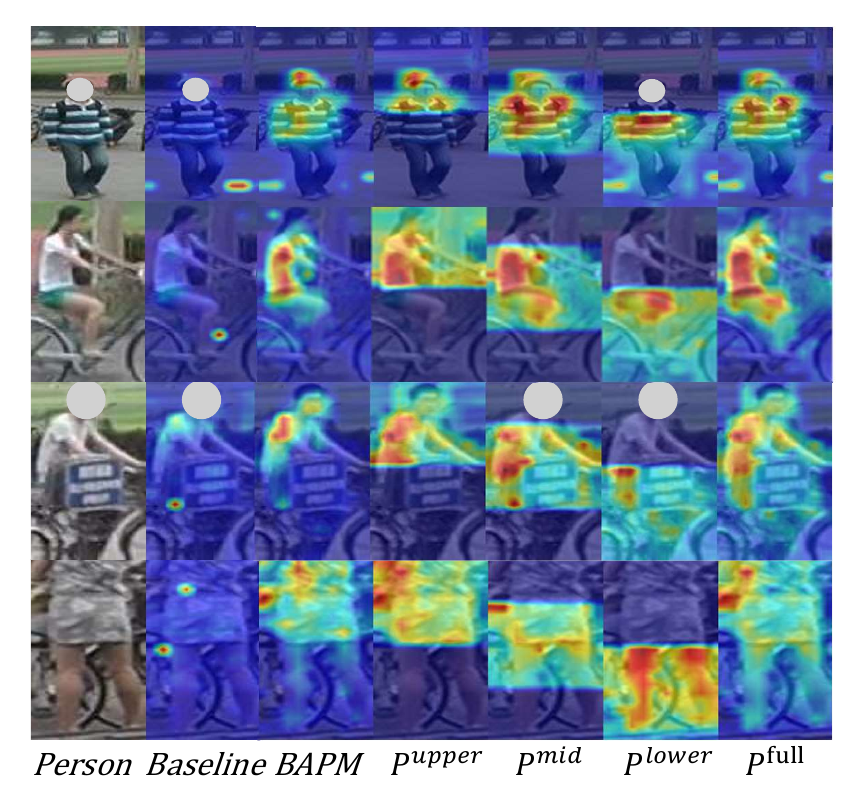}
    \vspace{-6mm}
    \caption{
        \textbf{Attention maps on Market-1501 under severe cropping, misalignment, and occlusion.}
        Unlike the scattered attention of the \textbf{Baseline}, our BAPM exhibits clear specialization: body part prompts ($\mathbf{P}^{\mathrm{upper}}$, \textit{etc.}) localize specific regions, while holistic prompts ($\mathbf{P}^{\mathrm{full}}$) capture global features. This aggregation yields comprehensive and accurate \textbf{BAPM} attention.
    }
    \label{fig:attention_maps}
\end{figure}

\begin{table}[th!]
\centering
\vspace{-2mm}
\begin{tabular}{c|c|c|c}
\Xhline{1.5pt}
\rowcolor{orange}
\multicolumn{2}{c|}{\textbf{Class Token}} & \multicolumn{2}{c}{\textbf{RISE}} \\
\hline
\rowcolor{orange}
\textbf{Method} & \textbf{Ins. AUC} & \textbf{Method} & \textbf{Ins. AUC} \\
\hline
\hline
SSCU & 0.6160 & SSCU & 0.6516 \\
\rowcolor{orangegray}
+VPs & 0.7103 & +VPs & 0.7494 \\
\rowcolor{mediumorangegray}
\textbf{BAPM} & \textbf{0.7559} & \textbf{BAPM} & \textbf{0.7737} \\
\Xhline{1.5pt}
\end{tabular}
\caption{\textbf{Comparison of Attention Map Types (Ins. AUC).} We evaluate the fidelity of different attention map generation methods using Insertion Area Under the Curve (Ins. AUC) metric.}
\vspace{-3  mm}
\label{tab:auc}
\end{table}

\subsection{Visualization}
\textbf{Visualization of attention maps.} We first visualize class token attention maps in Figure~\ref{fig:attention_maps}. The baseline model's attention is diffuse, scattering across the background. In contrast, our method concentrates class token attention on the person's body. To validate the effectiveness of our method in handling significant cross-client distribution shifts in background and viewpoint, we visualize samples under misalignment, cropping, and occlusion. Specifically, the part-alignment prompts $\mathbf{P}^{\mathrm{upper}}$, $\mathbf{P}^{\mathrm{mid}}$, and $\mathbf{P}^{\mathrm{lower}}$ correctly localize their designated body regions to ensure viewpoint-invariance, while the holistic prompt $\mathbf{P}^{\mathrm{full}}$ maintains focus on the entire person to suppress background interference.

To quantitatively validate this improved focus, we employ the Insertion AUC metric, which measures how rapidly performance is restored as critical pixels are revealed. As summarized in Table~\ref{tab:auc}, our method consistently achieves the highest Insertion AUC score compared against others. This holds true for attention maps derived from both the class token and RISE~\cite{rise}, confirming our method's superior ability to pinpoint crucial visual evidence.
\begin{figure}[h!]
    \centering
    \includegraphics[width=\linewidth]{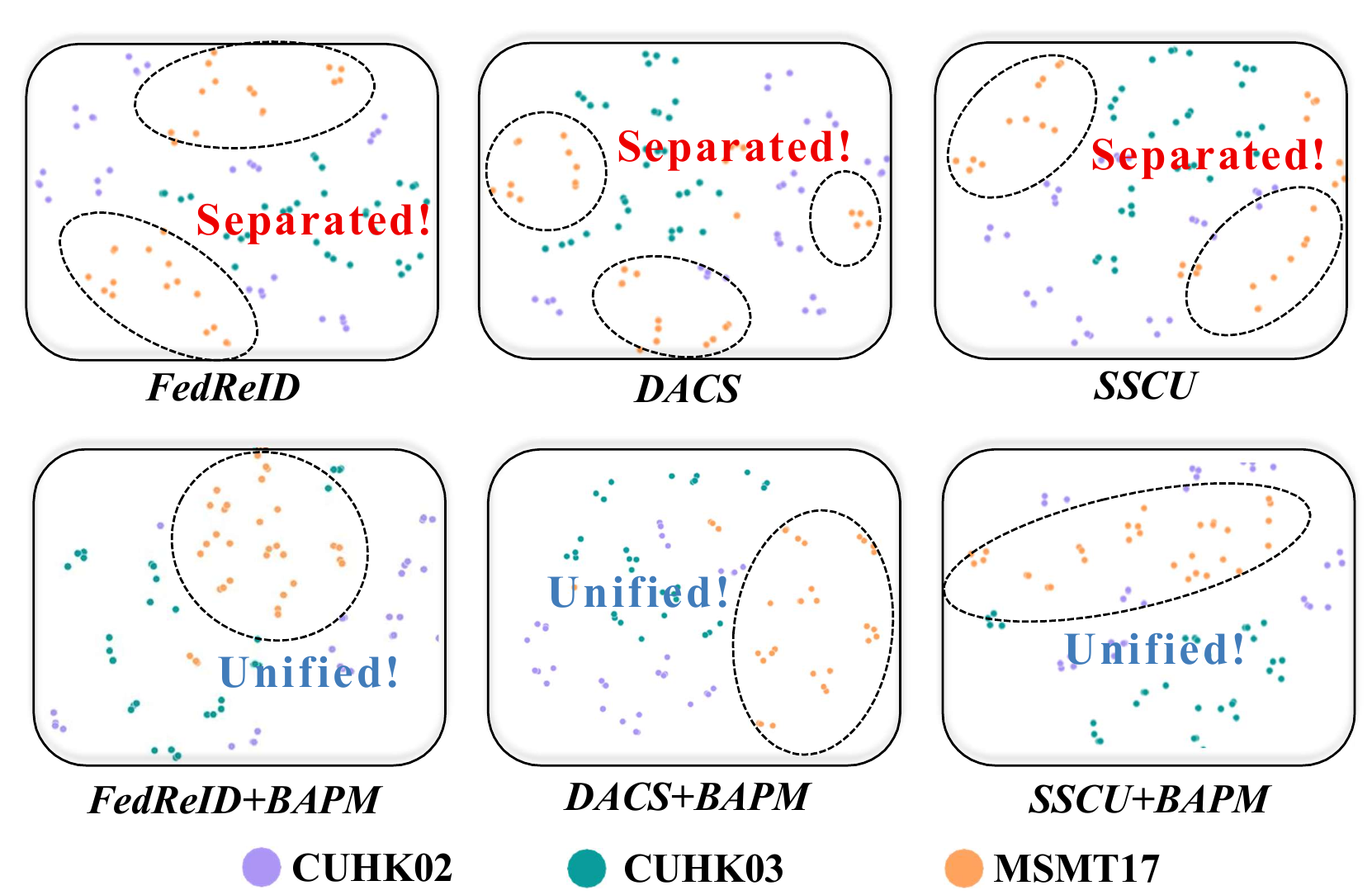}
    \caption{
        T-SNE visualizations of feature distribution on CUHK02 (C2), CUHK03 (C3), and MSMT17 (MS).
    }
    \vspace{-4mm}
    \label{fig:tsne}
\end{figure}

\textbf{Visualization of the Learned Feature Space.} Figure~\ref{fig:tsne}, visualizes the t-SNE embeddings of features from four baseline methods, compared with our BAPM (full-parameter), all under the ``MS+C2+C3'' setting. It is evident that both of our proposed methods yield feature distributions with significantly greater intra-domain compactness and inter-domain separability. This is particularly pronounced for the MSMT17 domain (circled). While the baseline methods scatter the MSMT17 samples, failing to form a cohesive cluster, our approach successfully aggregates them into a distinct and unified group.


\section{Conclusion}
We address the critical challenges in Federated Domain Generalization for Person Re-Identification (FedDG-ReID) that stem from significant \textbf{cross-client distribution shifts in background and viewpoint}. These shifts manifest as severe background interference and person misalignment, hindering the performance of existing models. To overcome this, we proposed \textbf{FedBPrompt}, a novel framework that leverages structured visual prompts to enhance generalization and robustness in a communication-efficient manner.
Our core contribution is the \textbf{Body Distribution Aware Visual Prompts Mechanism (BAPM)}, which is specifically designed to counter this heterogeneity. It partitions prompts into two functionally distinct groups: the Body Part Alignment Prompts learn viewpoint-invariant, part-level features to directly counter misalignment caused by diverse client viewpoints, while the Holistic Full Body Prompts maintain a robust focus on the entire pedestrian, mitigating distraction from the heterogeneous backgrounds. To address the high communication overhead, we introduced the \textbf{Prompt-based Fine-Tuning Strategy (PFTS)}, which freezes the backbone and exclusively communicates the lightweight prompts, reducing communication volume by over 99\%.


\section{Acknowledgments}
This work was supported by the Natural Science Foundation of China (62376201, 62501428, and 623B2080), Hubei Provincial Science \& Technology Talent Enterprise Services Program (2025DJB059), and Hubei Provincial Special Fund for Central-Guided Local S\&T Development (2025CSA017).

{
    \small
    \bibliographystyle{ieeenat_fullname}
    \bibliography{main}

\begin{thebibliography}{47}
\providecommand{\natexlab}[1]{#1}
\providecommand{\url}[1]{\texttt{#1}}
\expandafter\ifx\csname urlstyle\endcsname\relax
  \providecommand{\doi}[1]{doi: #1}\else
  \providecommand{\doi}{doi: \begingroup \urlstyle{rm}\Url}\fi

\bibitem[Bahng et~al.(2022)Bahng, Jahanian, Sankaranarayanan, and Isola]{bahng2022exploring}
Hyojin Bahng, Ali Jahanian, Swami Sankaranarayanan, and Phillip Isola.
\newblock Exploring visual prompts for adapting large-scale models.
\newblock \emph{arXiv preprint arXiv:2203.17274}, 2022.

\bibitem[Bugshan et~al.(2022)Bugshan, Khalil, Rahman, Atiquzzaman, Yi, and Badsha]{fl1}
Neda Bugshan, Ibrahim Khalil, Mohammad~Saidur Rahman, Mohammed Atiquzzaman, Xun Yi, and Shahriar Badsha.
\newblock Toward trustworthy and privacy-preserving federated deep learning service framework for industrial internet of things.
\newblock \emph{IEEE Transactions on Industrial Informatics}, 19\penalty0 (2):\penalty0 1535--1547, 2022.

\bibitem[Choi et~al.(2021)Choi, Kim, Jeong, Park, and Kim]{dgreid2}
Seokeon Choi, Taekyung Kim, Minki Jeong, Hyoungseob Park, and Changick Kim.
\newblock Meta batch-instance normalization for generalizable person re-identification.
\newblock In \emph{Proceedings of the IEEE/CVF conference on Computer Vision and Pattern Recognition}, pages 3425--3435, 2021.

\bibitem[Dai et~al.(2021)Dai, Li, Liu, Tong, and Duan]{dai2021generalizable}
Yongxing Dai, Xiaotong Li, Jun Liu, Zekun Tong, and Ling-Yu Duan.
\newblock Generalizable person re-identification with relevance-aware mixture of experts.
\newblock In \emph{Proceedings of the IEEE/CVF conference on computer vision and pattern recognition}, pages 16145--16154, 2021.

\bibitem[Dosovitskiy(2020)]{vit}
Alexey Dosovitskiy.
\newblock An image is worth 16x16 words: Transformers for image recognition at scale.
\newblock \emph{arXiv preprint arXiv:2010.11929}, 2020.

\bibitem[Gao et~al.(2022)Gao, Wei, Guan, Nie, Liu, and Wang]{reid1}
Zan Gao, Hongwei Wei, Weili Guan, Weizhi Nie, Meng Liu, and Meng Wang.
\newblock Multigranular visual-semantic embedding for cloth-changing person re-identification.
\newblock In \emph{Proceedings of the 30th ACM international conference on multimedia}, pages 3703--3711, 2022.

\bibitem[Guo et~al.(2023)Guo, Guo, Wang, Tang, and Xu]{promptfl}
Tao Guo, Song Guo, Junxiao Wang, Xueyang Tang, and Wenchao Xu.
\newblock Promptfl: Let federated participants cooperatively learn prompts instead of models--federated learning in age of foundation model.
\newblock \emph{IEEE Transactions on Mobile Computing}, 23\penalty0 (5):\penalty0 5179--5194, 2023.

\bibitem[Huang et~al.(2024)Huang, Ye, Shi, Wan, Li, Du, and Yang]{fl2}
Wenke Huang, Mang Ye, Zekun Shi, Guancheng Wan, He Li, Bo Du, and Qiang Yang.
\newblock Federated learning for generalization, robustness, fairness: A survey and benchmark.
\newblock \emph{IEEE Transactions on Pattern Analysis and Machine Intelligence}, 46\penalty0 (12):\penalty0 9387--9406, 2024.

\bibitem[Jia et~al.(2022)Jia, Tang, Chen, Cardie, Belongie, Hariharan, and Lim]{vpt}
Menglin Jia, Luming Tang, Bor-Chun Chen, Claire Cardie, Serge Belongie, Bharath Hariharan, and Ser-Nam Lim.
\newblock Visual prompt tuning.
\newblock In \emph{European conference on computer vision}, pages 709--727. Springer, 2022.

\bibitem[Jin et~al.(2020)Jin, Lan, Zeng, Chen, and Zhang]{snr}
Xin Jin, Cuiling Lan, Wenjun Zeng, Zhibo Chen, and Li Zhang.
\newblock Style normalization and restitution for generalizable person re-identification.
\newblock In \emph{Proceedings of the IEEE/CVF conference on computer vision and pattern recognition}, pages 3143--3152, 2020.

\bibitem[Karimireddy et~al.(2020)Karimireddy, Kale, Mohri, Reddi, Stich, and Suresh]{scaffold}
Sai~Praneeth Karimireddy, Satyen Kale, Mehryar Mohri, Sashank Reddi, Sebastian Stich, and Ananda~Theertha Suresh.
\newblock Scaffold: Stochastic controlled averaging for federated learning.
\newblock In \emph{International conference on machine learning}, pages 5132--5143. PMLR, 2020.

\bibitem[Kone{\v{c}}n{\`y} et~al.(2016)Kone{\v{c}}n{\`y}, McMahan, Yu, Richt{\'a}rik, Suresh, and Bacon]{fedavg}
Jakub Kone{\v{c}}n{\`y}, H~Brendan McMahan, Felix~X Yu, Peter Richt{\'a}rik, Ananda~Theertha Suresh, and Dave Bacon.
\newblock Federated learning: Strategies for improving communication efficiency.
\newblock \emph{arXiv preprint arXiv:1610.05492}, 2016.

\bibitem[Lester et~al.(2021)Lester, Al-Rfou, and Constant]{power_prompt_tuning}
Brian Lester, Rami Al-Rfou, and Noah Constant.
\newblock The power of scale for parameter-efficient prompt tuning.
\newblock \emph{arXiv preprint arXiv:2104.08691}, 2021.

\bibitem[Li et~al.(2021)Li, He, and Song]{moon}
Qinbin Li, Bingsheng He, and Dawn Song.
\newblock Model-contrastive federated learning.
\newblock In \emph{Proceedings of the IEEE/CVF conference on computer vision and pattern recognition}, pages 10713--10722, 2021.

\bibitem[Li et~al.(2020)Li, Sahu, Zaheer, Sanjabi, Talwalkar, and Smith]{fedprox}
Tian Li, Anit~Kumar Sahu, Manzil Zaheer, Maziar Sanjabi, Ameet Talwalkar, and Virginia Smith.
\newblock Federated optimization in heterogeneous networks.
\newblock \emph{Proceedings of Machine learning and systems}, 2:\penalty0 429--450, 2020.

\bibitem[Li and Wang(2013)]{cuhk02}
Wei Li and Xiaogang Wang.
\newblock Locally aligned feature transforms across views.
\newblock In \emph{Proceedings of the IEEE conference on computer vision and pattern recognition}, pages 3594--3601, 2013.

\bibitem[Li et~al.(2014)Li, Zhao, Xiao, and Wang]{cuhk03}
Wei Li, Rui Zhao, Tong Xiao, and Xiaogang Wang.
\newblock Deepreid: Deep filter pairing neural network for person re-identification.
\newblock In \emph{Proceedings of the IEEE conference on computer vision and pattern recognition}, pages 152--159, 2014.

\bibitem[Li and Liang(2021)]{li2021prefix}
Xiang~Lisa Li and Percy Liang.
\newblock Prefix-tuning: Optimizing continuous prompts for generation.
\newblock \emph{arXiv preprint arXiv:2101.00190}, 2021.

\bibitem[Liu et~al.(2024)Liu, Ye, and Du]{liu2024domain}
Fangyi Liu, Mang Ye, and Bo Du.
\newblock Domain generalized federated learning for person re-identification.
\newblock \emph{Computer Vision and Image Understanding}, 241:\penalty0 103969, 2024.

\bibitem[Liu et~al.(2023)Liu, Yuan, Fu, Jiang, Hayashi, and Neubig]{prompt}
Pengfei Liu, Weizhe Yuan, Jinlan Fu, Zhengbao Jiang, Hiroaki Hayashi, and Graham Neubig.
\newblock Pre-train, prompt, and predict: A systematic survey of prompting methods in natural language processing.
\newblock \emph{ACM computing surveys}, 55\penalty0 (9):\penalty0 1--35, 2023.

\bibitem[Liu et~al.(2021{\natexlab{a}})Liu, Chen, Qin, Dou, and Heng]{feddg1}
Quande Liu, Cheng Chen, Jing Qin, Qi Dou, and Pheng-Ann Heng.
\newblock Feddg: Federated domain generalization on medical image segmentation via episodic learning in continuous frequency space.
\newblock In \emph{Proceedings of the IEEE/CVF conference on computer vision and pattern recognition}, pages 1013--1023, 2021{\natexlab{a}}.

\bibitem[Liu et~al.(2021{\natexlab{b}})Liu, Ji, Fu, Tam, Du, Yang, and Tang]{P-Tuning_v2}
Xiao Liu, Kaixuan Ji, Yicheng Fu, Weng~Lam Tam, Zhengxiao Du, Zhilin Yang, and Jie Tang.
\newblock P-tuning v2: Prompt tuning can be comparable to fine-tuning universally across scales and tasks.
\newblock \emph{arXiv preprint arXiv:2110.07602}, 2021{\natexlab{b}}.

\bibitem[Luo et~al.(2019)Luo, Gu, Liao, Lai, and Jiang]{luo}
Hao Luo, Youzhi Gu, Xingyu Liao, Shenqi Lai, and Wei Jiang.
\newblock Bag of tricks and a strong baseline for deep person re-identification.
\newblock In \emph{Proceedings of the IEEE/CVF conference on computer vision and pattern recognition workshops}, pages 0--0, 2019.

\bibitem[Nguyen et~al.(2024)Nguyen, Nguyen, Sridharan, and Fookes]{reid7}
Huy Nguyen, Kien Nguyen, Sridha Sridharan, and Clinton Fookes.
\newblock Ag-reid. v2: Bridging aerial and ground views for person re-identification.
\newblock \emph{IEEE Transactions on Information Forensics and Security}, 19:\penalty0 2896--2908, 2024.

\bibitem[Ni et~al.(2022)Ni, Song, Luo, Zheng, Li, and Shen]{ni2022meta}
Hao Ni, Jingkuan Song, Xiaopeng Luo, Feng Zheng, Wen Li, and Heng~Tao Shen.
\newblock Meta distribution alignment for generalizable person re-identification.
\newblock In \emph{Proceedings of the IEEE/CVF conference on computer vision and pattern recognition}, pages 2487--2496, 2022.

\bibitem[Ni et~al.(2023)Ni, Li, Gao, Shen, and Song]{ni2023part}
Hao Ni, Yuke Li, Lianli Gao, Heng~Tao Shen, and Jingkuan Song.
\newblock Part-aware transformer for generalizable person re-identification.
\newblock In \emph{Proceedings of the IEEE/CVF international conference on computer vision}, pages 11280--11289, 2023.

\bibitem[Petsiuk et~al.(2018)Petsiuk, Das, and Saenko]{rise}
Vitali Petsiuk, Abir Das, and Kate Saenko.
\newblock Rise: Randomized input sampling for explanation of black-box models.
\newblock \emph{arXiv preprint arXiv:1806.07421}, 2018.

\bibitem[Su et~al.(2017)Su, Li, Zhang, Xing, Gao, and Tian]{reid4}
Chi Su, Jianing Li, Shiliang Zhang, Junliang Xing, Wen Gao, and Qi Tian.
\newblock Pose-driven deep convolutional model for person re-identification.
\newblock In \emph{Proceedings of the IEEE international conference on computer vision}, pages 3960--3969, 2017.

\bibitem[Sun et~al.(2018)Sun, Zheng, Yang, Tian, and Wang]{pcb}
Yifan Sun, Liang Zheng, Yi Yang, Qi Tian, and Shengjin Wang.
\newblock Beyond part models: Person retrieval with refined part pooling (and a strong convolutional baseline).
\newblock In \emph{Proceedings of the European conference on computer vision (ECCV)}, pages 480--496, 2018.

\bibitem[Tang et~al.(2021)Tang, Gao, Zhu, Zhang, Li, and Metaxas]{crossstyle}
Zhiqiang Tang, Yunhe Gao, Yi Zhu, Zhi Zhang, Mu Li, and Dimitris~N Metaxas.
\newblock Crossnorm and selfnorm for generalization under distribution shifts.
\newblock In \emph{Proceedings of the IEEE/CVF International Conference on Computer Vision}, pages 52--61, 2021.

\bibitem[Wang et~al.(2018)Wang, Yuan, Chen, Li, and Zhou]{mgn}
Guanshuo Wang, Yufeng Yuan, Xiong Chen, Jiwei Li, and Xi Zhou.
\newblock Learning discriminative features with multiple granularities for person re-identification.
\newblock In \emph{Proceedings of the 26th ACM international conference on Multimedia}, pages 274--282, 2018.

\bibitem[Wei et~al.(2018)Wei, Zhang, Gao, and Tian]{msmt17}
Longhui Wei, Shiliang Zhang, Wen Gao, and Qi Tian.
\newblock Person transfer gan to bridge domain gap for person re-identification.
\newblock In \emph{Proceedings of the IEEE conference on computer vision and pattern recognition}, pages 79--88, 2018.

\bibitem[Wu and Gong(2021)]{wu2021decentralised}
Guile Wu and Shaogang Gong.
\newblock Decentralised learning from independent multi-domain labels for person re-identification.
\newblock In \emph{Proceedings of the AAAI Conference on Artificial Intelligence}, pages 2898--2906, 2021.

\bibitem[Xu et~al.(2022)Xu, Liu, Wang, Hu, and Tian]{wei}
Xin Xu, Wei Liu, Zheng Wang, Ruimin Hu, and Qi Tian.
\newblock Towards generalizable person re-identification with a bi-stream generative model.
\newblock \emph{Pattern Recognition}, 132:\penalty0 108954, 2022.

\bibitem[Xu et~al.(2025)Xu, Ren, Liu, Huang, Yang, Yu, and Jiang]{sscu}
Xin Xu, Chaoyue Ren, Wei Liu, Wenke Huang, Bin Yang, Zhixi Yu, and Kui Jiang.
\newblock Positive style accumulation: A style screening and continuous utilization framework for federated dg-reid.
\newblock In \emph{Proceedings of the 33rd ACM International Conference on Multimedia}, pages 8527--8536, 2025.

\bibitem[Xu et~al.(2026)Xu, Ma, Yu, and Liu]{xu2026fedarks}
Xin Xu, Binchang Ma, Zhixi Yu, and Wei Liu.
\newblock Fedarks: Federated aggregation via robust and discriminative knowledge selection and integration for person re-identification.
\newblock \emph{arXiv preprint arXiv:2603.06122}, 2026.

\bibitem[Yan et~al.(2021)Yan, Hao, Li, Yin, Liu, Mao, Chen, and Gao]{reid2}
Chenggang Yan, Yiming Hao, Liang Li, Jian Yin, Anan Liu, Zhendong Mao, Zhenyu Chen, and Xingyu Gao.
\newblock Task-adaptive attention for image captioning.
\newblock \emph{IEEE Transactions on Circuits and Systems for Video technology}, 32\penalty0 (1):\penalty0 43--51, 2021.

\bibitem[Yan et~al.(2022)Yan, Meng, Li, Zhang, Wang, Yin, Zhang, Sun, and Zheng]{reid5}
Chenggang Yan, Lixuan Meng, Liang Li, Jiehua Zhang, Zhan Wang, Jian Yin, Jiyong Zhang, Yaoqi Sun, and Bolun Zheng.
\newblock Age-invariant face recognition by multi-feature fusionand decomposition with self-attention.
\newblock \emph{ACM Transactions on Multimedia Computing, Communications, and Applications (TOMM)}, 18\penalty0 (1s):\penalty0 1--18, 2022.

\bibitem[Yang et~al.(2022)Yang, Zhong, Luo, Li, and Sebe]{yang2022federated}
Fengxiang Yang, Zhun Zhong, Zhiming Luo, Shaozi Li, and Nicu Sebe.
\newblock Federated and generalized person re-identification through domain and feature hallucinating.
\newblock \emph{arXiv preprint arXiv:2203.02689}, 2022.

\bibitem[Yang et~al.(2024)Yang, Zhong, Luo, He, Li, and Sebe]{dacs}
Fengxiang Yang, Zhun Zhong, Zhiming Luo, Yifan He, Shaozi Li, and Nicu Sebe.
\newblock Diversity-authenticity co-constrained stylization for federated domain generalization in person re-identification.
\newblock In \emph{Proceedings of the AAAI Conference on Artificial Intelligence}, pages 6477--6485, 2024.

\bibitem[Zhang et~al.(2023)Zhang, Xu, Yao, Zhang, Tian, and Wang]{feddg2}
Ruipeng Zhang, Qinwei Xu, Jiangchao Yao, Ya Zhang, Qi Tian, and Yanfeng Wang.
\newblock Federated domain generalization with generalization adjustment.
\newblock In \emph{Proceedings of the IEEE/CVF Conference on Computer Vision and Pattern Recognition}, pages 3954--3963, 2023.

\bibitem[Zhang et~al.(2022)Zhang, Zhang, Li, Jia, Wang, and Tan]{zhang2022learning}
Yi-Fan Zhang, Zhang Zhang, Da Li, Zhen Jia, Liang Wang, and Tieniu Tan.
\newblock Learning domain invariant representations for generalizable person re-identification.
\newblock \emph{IEEE Transactions on Image Processing}, 32:\penalty0 509--523, 2022.

\bibitem[Zhao et~al.(2023)Zhao, Du, Li, Li, and Liu]{fedprompt}
Haodong Zhao, Wei Du, Fangqi Li, Peixuan Li, and Gongshen Liu.
\newblock Fedprompt: Communication-efficient and privacy-preserving prompt tuning in federated learning.
\newblock In \emph{ICASSP 2023-2023 IEEE International Conference on Acoustics, Speech and Signal Processing (ICASSP)}, pages 1--5. IEEE, 2023.

\bibitem[Zhao et~al.(2021)Zhao, Zhong, Yang, Luo, Lin, Li, and Sebe]{dgreid1}
Yuyang Zhao, Zhun Zhong, Fengxiang Yang, Zhiming Luo, Yaojin Lin, Shaozi Li, and Nicu Sebe.
\newblock Learning to generalize unseen domains via memory-based multi-source meta-learning for person re-identification.
\newblock In \emph{Proceedings of the IEEE/CVF conference on computer vision and pattern recognition}, pages 6277--6286, 2021.

\bibitem[Zheng et~al.(2015)Zheng, Shen, Tian, Wang, Wang, and Tian]{market1501}
Liang Zheng, Liyue Shen, Lu Tian, Shengjin Wang, Jingdong Wang, and Qi Tian.
\newblock Scalable person re-identification: A benchmark.
\newblock In \emph{Proceedings of the IEEE international conference on computer vision}, pages 1116--1124, 2015.

\bibitem[Zhou et~al.(2021)Zhou, Yang, Qiao, and Xiang]{mixstyle}
Kaiyang Zhou, Yongxin Yang, Yu Qiao, and Tao Xiang.
\newblock Domain generalization with mixstyle.
\newblock \emph{arXiv preprint arXiv:2104.02008}, 2021.

\bibitem[Zhuang et~al.(2020)Zhuang, Wen, Zhang, Gan, Yin, Zhou, Zhang, and Yi]{fedpav}
Weiming Zhuang, Yonggang Wen, Xuesen Zhang, Xin Gan, Daiying Yin, Dongzhan Zhou, Shuai Zhang, and Shuai Yi.
\newblock Performance optimization of federated person re-identification via benchmark analysis.
\newblock In \emph{Proceedings of the 28th ACM international conference on multimedia}, pages 955--963, 2020.

\end{thebibliography}
}


\end{document}